\documentclass[10pt,twocolumn,letterpaper]{article}

 \usepackage{cvpr}              %

\usepackage{amsmath}
\usepackage{mathtools}
\usepackage{overpic}
\usepackage{color}
\usepackage{listings} %
\usepackage[skip=.2em]{caption} %
\usepackage{xcolor}
\usepackage{currfile}
\usepackage{cancel}
\usepackage{float}
\usepackage{rotating}
\usepackage{booktabs}
\usepackage{comment}

\usepackage{tabularx}
\usepackage{etoolbox}

\usepackage{cuted} %

\usepackage{currfile} %

\usepackage{soul}
\setuldepth{foobar}

\usepackage{caption} %

\usepackage[rightcaption]{sidecap}  %
\sidecaptionvpos{figure}{t} %
\definecolor{turquoise}{cmyk}{0.65,0,0.1,0.3}
\definecolor{purple}{rgb}{0.65,0,0.65}
\definecolor{dark_green}{rgb}{0, 0.5, 0}
\definecolor{orange}{rgb}{0.8, 0.6, 0.2}
\definecolor{red}{rgb}{0.8, 0.2, 0.2}
\definecolor{darkgray}{rgb}{0.5, 0.5, 0.5}
\definecolor{darkred}{rgb}{0.6, 0.1, 0.05}
\definecolor{blueish}{rgb}{0.0, 0.3, .6}
\definecolor{light_gray}{rgb}{0.7, 0.7, .7}
\definecolor{pink}{rgb}{1, 0, 1}
\definecolor{greyblue}{rgb}{0.25, 0.25, 1}

\usepackage{lipsum}
\usepackage{blindtext}

\renewcommand{\paragraph}[1]{\vspace{.25em}\noindent\textbf{#1}.}

\newcommand{\benchmark}{ORBIT\xspace}

\newcommand{\SupplementaryMaterial}[1]{{supplementary material}}

\usepackage{enumitem}
\setlist[itemize]{noitemsep,leftmargin=*,topsep=0em}
\setlist[enumerate]{noitemsep,leftmargin=*,topsep=0em}

\newcommand{\ATE}{\mathrm{ATE}}

\newcommand{\RPEt}{\mathrm{RPE_t}}
\newcommand{\RPEr}{\mathrm{RPE_r}}

\definecolor{cvprblue}{rgb}{0.21,0.49,0.74}
\usepackage[pagebackref,breaklinks,colorlinks,allcolors=cvprblue]{hyperref}

\def\paperID{00000} %
\def\confName{CVPR}
\def\confYear{2026}

\title{ORBIT++: Benchmarking SfM in the Wild with 360° Video}

\author{Sara Sabour\textsuperscript{1} \quad
Linyi Jin\textsuperscript{1} \quad
Richard Tucker\textsuperscript{1} \quad
Amir Hertz\textsuperscript{1} \quad
Marcus Brubaker\textsuperscript{1} \\
Saurabh Saxena\textsuperscript{1} \quad
Junhwa Hur\textsuperscript{1} \quad
Andrea Tagliasacchi\textsuperscript{3} \quad
Deqing Sun\textsuperscript{1} \quad
David J. Fleet\textsuperscript{1,2} \\
Richard Szeliski\textsuperscript{1} \quad
Noah Snavely\textsuperscript{1} \\
\small
\textsuperscript{1}Google DeepMind
\quad
\textsuperscript{2}University of Toronto
\quad
\textsuperscript{3}Simon Fraser University \\
\texttt{\small sasabour@, snavely@google.com}
}

\begin{document}
\maketitle
\begin{strip}
\vspace*{-1.0cm}
\begin{center}
\includegraphics[width=0.975\linewidth]{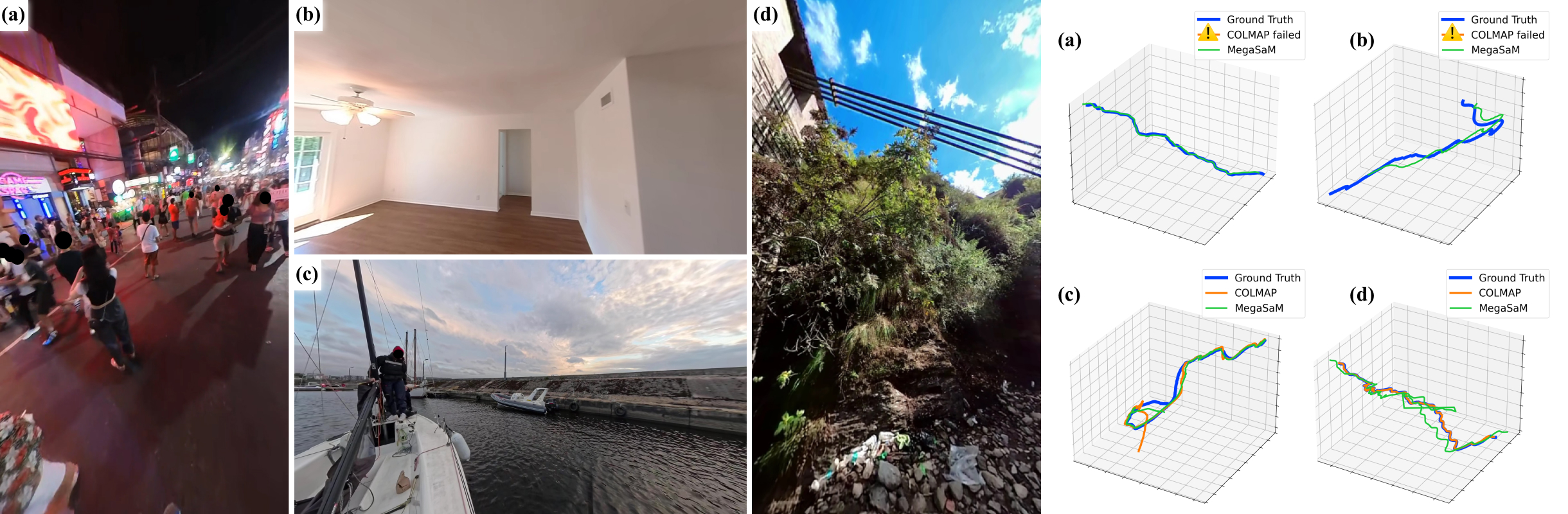}
\end{center}
\vspace*{-0.2em}
\captionof{figure}{\textbf{The ORBIT Structure-from-Motion dataset}. ORBIT includes a variety of clips that are challenging for SfM methods. Scenes with many dynamic objects~(a) or large textureless areas~(b) are challenging for traditional approaches like COLMAP. The ML-based MegaSaM handles these examples but can struggle when depth-estimation is difficult~(d). Bodies of water and large moving foreground objects such as vehicles~(c) present difficulties for both methods. [Here, ``Failed" means a method tracked fewer than 10 frames of the clip.]
}
\label{fig:\currfilebase}
\end{strip}

\begin{abstract}
\vspace*{-0.83cm}

Structure-from-Motion (SfM) is a cornerstone of 3D perception, yet current methods often fail when applied to complex videos involving challenging camera motions or dynamic scenes.
Compounding the problem, the field lacks reliable ground-truth benchmarks for such difficult scenarios, making it hard to gauge real-world progress or to pinpoint where improvements are most needed.
To address this gap, we introduce a new benchmark for evaluating camera pose estimation.
Our key insight is to leverage online panoramic 360° video as a source of data from which to construct challenging clips, while still enabling robust ground-truth trajectory recovery.
The panoramic nature of these videos provides richer visual context for tracking camera motion, even when parts of the view are affected by blur, motion, or dynamic objects.
After tracking camera motion across full 360° videos, we crop and reproject selected portions to generate perspective-view clips that serve as our benchmark, called ORBIT.
Experiments show that COLMAP, as well as recent optimization-based and feed-forward 
SfM methods struggle to accurately estimate camera 
poses on our benchmark.
Hence, ORBIT provides a valuable testbed where researchers can meaningfully measure progress on truly challenging, real-world SfM problems. The benchmark and related codebase are available at 
\url{https://orbit-sfm.github.io}.\footnote{
This work is an updated version of our CVPR 2026 conference paper that presents the larger dataset released on our website and more in-depth analysis and descriptions.
}
\end{abstract}

\section{Introduction}
\vspace*{-0.1cm}

Recovering 3D information from 2D
videos is key to spatial reasoning, visual prediction, fine-grained robotic interaction, AR/VR, and learning controllable (3D and 4D) generative world models. 
A central component of 3D recovery is 
structure-from-motion (SfM), where we take videos (or images) and produce camera poses and 3D scene geometry.

Current SfM methods work well on short videos featuring static scenes, but, even with significant recent progress, current methods can still produce errors or fail entirely on complex dynamic scenes---for instance, scenes featuring significant numbers of moving objects, regions of complex motion (e.g., waving foliage or moving water), and complex reflectance from specular surfaces; see \Cref{fig:teaser}.
Furthermore, it is difficult to gauge the performance of SfM methods on complex real-world videos, as benchmarks with ground truth camera poses that reflect these challenges are sorely missing.

Unfortunately, it is  very difficult to obtain high-quality ground truth camera trajectories and 3D information for challenging real-world videos. 
Much of the prior work uses COLMAP~\cite{schonberger2016colmap} as a source of truth, while in reality, we observe that COLMAP fails on \textit{many} real-world videos.
Gathering ground truth camera poses is possible in restricted settings, e.g., with GPS, IMUs, depth sensors, or instrumented scenes~\cite{kayan2025princeton365}, but such extra information is rarely available for in-the-wild data.

In this paper, we build such a benchmark by leveraging the growing availability of a special class of videos: 
panoramic videos captured by 360° cameras such as the Ricoh Theta or Insta360. 
For each frame, such devices capture a full sphere of rays around the camera center.
Our insight is that, for such videos, structure-from-motion is much more reliable than for standard narrow field of view videos.
First, even if much of the scene contains difficult content~(e.g., dynamic elements), other parts of the scene are likely to be static and therefore provide good features for camera motion estimation---in other words, stable features cannot ``hide'' from the camera, since the camera \textit{sees in all directions}.
Second, 360° videos have \textit{known intrinsics}, unlike regular in-the-wild videos where focal length may be unknown a priori.
Finally, pose estimation problems are fundamentally better constrained when working with wide field-of-view imagery~\cite{levin2004visualodometry,levin2006motionuncertainty}.

To build this new benchmark, 
we first \textit{curate} a diverse set of 360° videos from the web, featuring moving cameras and difficult scene content~(moving objects, water, distractor objects fixed to the camera frame, and so on).
We then run a custom SfM method designed 
for 360° videos, 
which yields reliable camera poses that we use as our pseudo-ground truth.
To form a difficult set of benchmark videos, we \textit{project} each 360° video to a perspective video that focuses on a difficult part of the scene, as measured by whether our baseline SfM method fails to accurately reproduce the pseudo-ground truth on that perspective crop.
We assemble a set of 300 evaluation videos using this process, each up to roughly 30 seconds in length, 
to form our benchmark, which we call \textbf{\benchmark}, the \textbf{O}mnidirectional \textbf{R}econstruction \textbf{B}enchmark with \textbf{I}mage \textbf{T}rajectories.

We use \benchmark to evaluate a number of classic and recent SfM methods, including COLMAP~\cite{schonberger2016colmap}, MegaSaM~\cite{li2025megasam}, and VGGT~\cite{wang2025vggt}.
Our analysis shows that there is significant room for improvement among even such state-of-the-art techniques.
Every method completely fails to estimate camera poses on \textit{at least} 36\% of clips.  
More importantly, the different types of challenges we categorize enable the analysis and comparison of failure modes between methods.
Therefore, we find that \benchmark is a strong tool for diagnosing and evaluating current SfM methods, paving the way towards future improvements.

\section{Related Work}
\vspace*{-0.1cm}

Traditional camera pose estimation methods such as COLMAP~\cite{schonberger2016colmap} can reliably and accurately register a set of unordered images into a 3D reconstruction using a well-engineered suite of handcrafted methods such as feature matching and incremental pose and structure estimation. 
If the images are ordered in time, such as in a video, Simultaneous Localization and Mapping (SLAM) methods like ORB-SLAM~\cite{mur2015orb} can use camera tracking techniques and incrementally build a 3D map, increasing efficiency to real time.
However, 
these classic methods lack learned priors to guide them in 
challenging cases that arise in in-the-wild capture such as textureless scenes or 
dynamic objects.

\paragraph{Learning-based methods}
To overcome these challenges,
more recent approaches have incorporated learnable modules. Notable examples that adopt an optimization paradigm include DROID-SLAM~\cite{Teed2021droidslam}, CasualSAM~\cite{zhang2022structure}, and MegaSaM~\cite{li2025megasam}.
To address specific challenges posed by dynamic objects, one popular approach is to incorporate motion segmentation within the process, as in 
ParticleSFM~\cite{zhao2022particlesfm} or RoMo~\cite{goli2025romo}.
Recently, feed-forward methods such as VGGT~\cite{wang2025vggt} and MonST3R~\cite{zhang2024monst3r} 
learn to regress directly from images to scene reconstructions and camera poses
by training on large annotated datasets like those discussed below.

\begin{figure*}[ht]
\begin{center}
\includegraphics[width=\linewidth]{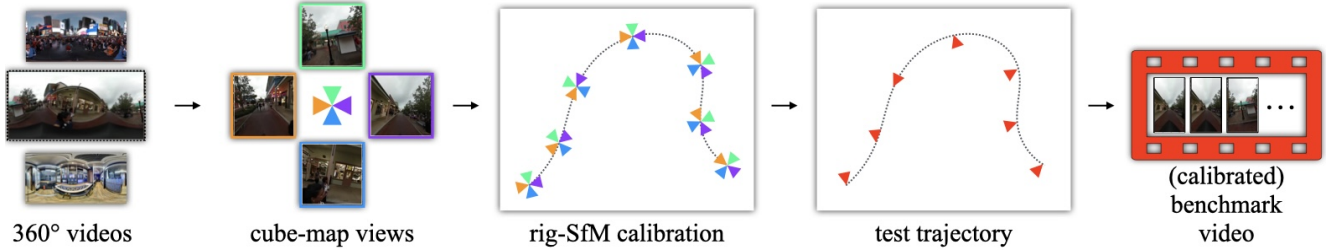}
\end{center}
\vspace*{-.3em}
\captionof{figure}{
\textbf{Pipeline.}
We find suitable 360° videos on the Internet, convert each video into cube-maps to estimate their poses with our rig-based SfM algorithm, and then select a test trajectory (time-varying viewing directions) from which each  benchmark video is rendered.
}
\label{fig:\currfilebase}
\end{figure*}

\paragraph{Datasets}
Modern large models trained for 3D tasks like novel view synthesis or 3D reconstruction 
still rely on traditional pipelines such as COLMAP to annotate data with camera poses and 3D structure.
Unfortunately, classical SfM and SLAM systems are typically evaluated on \textit{synthetic} datasets (e.g., Sintel~\cite{butler2012sintel} or TartanAir~\cite{wang2020tartanair}).
Many methods focus on \textit{static} scenes from datasets such as RealEstate10k~\cite{zhou2018stereo} or ETH3D~\cite{schops2019badslam}, where camera trajectories are relatively simple and do not reflect the complexity of real-world motion.
In contrast, datasets like DyCheck~\cite{gao2022monocular}, Waymo~\cite{Sun2020waymo}, and city-scale benchmarks~\cite{lindenberger2025scaling} include natural camera trajectories but exhibit limited scene diversity due to their narrow domain coverage.

The absence of comprehensive benchmarks and training datasets for dynamic, real-world videos, where both the camera and objects move, has motivated several recent efforts to curate such resources.
DynPose-100K~\cite{rockwell2025dynpose} and SpatialVID~\cite{wang2025spatialvid} may seem adequate, as they provide videos with estimated camera positions.
Nevertheless, these annotations are \textit{predicted} using pose estimation systems such as COLMAP and MegaSaM~\cite{li2025megasam}, making their accuracy difficult to verify and thus unsuitable for SfM benchmarking. DynPose-100K relies on the same monocular test inputs for its camera estimates, as opposed to our ORBIT benchmark, which uses different information for verification and test trajectories. 
In particular, we 
compute ground truth camera poses from 360° videos by leveraging the full field of view, and  
generate evaluation videos by projecting to perspective 
projections subtending a limited 
field of view. 
As such, DynPose-100K, due to lack of reliability in its ground truth estimates, 
is only reliable as a (noisy) training dataset (e.g., for finetuning models like DUSt3R) whereas ORBIT is suitable as challenging benchmark dataset.

\paragraph{360° datasets}
ORBIT is constructed from a carefully curated collection of 360° videos sourced from the web,
from which perspective camera trajectories and positions are derived. 
Princeton365 \cite{kayan2025princeton365} is another recent 360°video--based dataset. Princeton365 uses IMUs/markers to calibrate their ground truth sequences, limiting it to more controlled setups in constrained environments; it lacks the ``in-the-wild'' diversity of ORBIT (e.g., canoeing, crowds). 
\citet{wallingford2024fromanimage} also utilize 360° videos, but their focus on training generative models differs from ours. 
In our work, ORBIT serves as a benchmark designed to challenge SfM and SLAM methods, necessitating a diverse and rigorously filtered selection of videos, for which the web provides a diverse and extensive data pool.

\section{Building \benchmark}
\label{sec:method}

\vspace*{-0.1cm}
We extract our benchmark videos from online 360° panoramic videos.
Such videos are plentiful on the web due to the popularity of cameras like the Insta360 and Ricoh Theta.
First, we identify a challenging, diverse, set of 360° online video clips to form the basis of our camera pose estimation benchmark~(\Cref{sec:filtering}). 
Second, we estimate camera poses from these videos with an SfM pipeline customized for 360° video, and perform a cross-validation procedure to assess their viability as ground truth camera poses~(\Cref{sec:annotation}). 
Finally, we reproject these 360° videos to form challenging, interesting perspective videos, with a variety of fields of view and rotational behaviors, while inheriting ground truth  estimates of camera position from the original 360° videos~(\Cref{sec:synthesizing}).
We depict the process we use to create ORBIT in~\Cref{fig:pipeline}.

\paragraph{Summary statistics}
After applying all of these pose estimation, filtering, and cropping stages, \benchmark comprises 308  clips of various lengths.
The clips are selected from 77 unique 360° videos and have 150 to 1,000 frames with a frame rate of 30 fps. 
The clips have either an aspect ratio of $16{:}9$ or $4{:}3$ in portrait or landscape mode, as these are typical of consumer captures, with resolutions varying between  $ 384  \times 512$ and $360 \times 640$, accompanied by ground truth camera trajectories.

\subsection{Identifying suitable videos}
\label{sec:filtering}
We first identify online 360° videos with a 1K minimum resolution, so that after reprojection into our desired perspective field-of-view videos we can attain at least VGA~(480p) compatible resolution.
We then look for videos that meet camera motion and camera continuity requirements.
In particular, we find that many online 360° videos are either a concatenation of static images, a concatenation of short clips, or a completely static camera observing a scene.
We manually select videos that feature a single camera moving across a long, continuous shot, and also identify videos whose content is likely challenging for camera pose estimation, including videos characterized by difficult environments (e.g., snow, water, sand, variable lighting, fast camera speeds, or scene dynamics).
We then break the videos into shorter 2,000-frame clips without overlap to accommodate the limitations of current SFM and SLAM methods.

\begin{figure*}[ht]
\begin{center}
\includegraphics[width=0.9\linewidth]{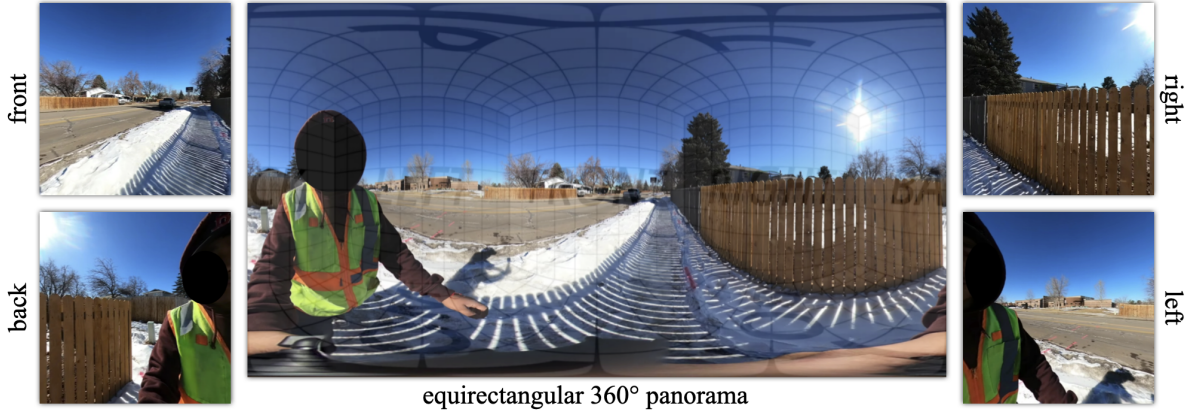}
\end{center}
\vspace*{-0.3cm}
\captionof{figure}{
\textbf{Reprojected cube face images.}
A 360° panoramic video frame (middle) is projected to four perspective images corresponding to four faces of a cube (front, right, back, and left) centered at the camera's center of projection,
since, as mentioned in the text the equirectangular projection is \textit{highly anisotropic}, making feature matching and tracking more challenging.
}
\label{fig:\currfilebase}
\end{figure*}

\subsection{Ground truth trajectory estimation}
\label{sec:annotation}
Given each candidate video, we obtain high-quality camera pose estimates (that serve as our ground truth once validated)
by estimating robust camera poses with a specially designed 360° \textit{rig-based SfM} procedure (\cref{sec:bundle}).
Given these camera poses, initially at an arbitrary scale, we convert them to approximately metric scale in order to make metrics, such as translation error, consistent across all videos (\cref{sec:metricscale}).
We then validate our ground truth estimates by cross-validating SfM and SLAM camera poses on different projections of the video, dropping clips where cross-validation fails to support the ground truth proposed by the full 360° video (\cref{sec:gtverify}).

\subsubsection{Rig-based pose estimation}
\label{sec:bundle}
Given a 360° video, we estimate per-frame 3D camera poses, along with a 3D point cloud, in a way that makes joint use of all information available in the input. 
The most common format for online 360° videos is an \emph{equirectangular} format, a spherical projection of the video where the position of a pixel in each frame depends on the latitude and longitude of the ray 
through that pixel when 
intersected with a unit sphere centered on the camera position.
A number of methods for SfM/SLAM from 360° imagery have been proposed in the past.
One simple strategy for handling such videos is to adapt a standard SfM method like COLMAP to a new kind of camera model representing an equirectangular projection.
Such an approach involves a few challenges:
\begin{enumerate*}[label=(\roman*)]
\item the equirectangular projection is \textit{highly anisotropic}, in that a small pixel offset at the equator of the projection corresponds to a much larger offset between 3D rays compared to the same offset near the poles; hence, reprojection error needs to be computed with care;
\item the \textit{projection wraps} (the left and right sides of the projection are connected), so 3D points that project near the boundary must be handled specially.
\end{enumerate*}

\paragraph{Cube-face sub-videos}
To avoid such difficulties, we take a different approach.  We treat each 360° video frame as having been captured by a rigidly-attached rig of multiple perspective cameras, where each camera corresponds to a perspective image projected from the raw equirectangular frame from a specific orientation.
Conceptually, one can think of reprojecting the equirectangular (spherical) projection to a cube map projection (six faces of a cube surrounding the center), and treating each cube face as an image captured by a perspective camera, rigidly attached to the cameras viewing the other faces.
This approach is inspired by prior work that treats panoramic reconstruction as handling images captured from separate cameras~\cite{uyttendaele2004imagebased, kangni2010orientation, klingner2013streetview}.
Given a video, we project each equirectangular frame to four cube faces (front, back, left, right in~\Cref{fig:projection}).
We call these ``cube faces'', although, for added robustness, we use a 120° field of view, rather than 90°, meaning that each face \textit{overlaps} its neighbors.
We omit the top and bottom cube faces since they often contain missing or spurious information, e.g., a graphic is often pasted at the south pole of the video, while the north pole is often pointed at a featureless sky.
We therefore consider the original video to be broken into {four} temporally {synchronized} {perspective} sub-videos.

\paragraph{Estimating the rig pose}
We run ORB-SLAM2~\cite{mur2015orb}, a standard SLAM pipeline, on the front-facing cube face as a way of bootstrapping the camera geometry for the video as a whole, since ORB-SLAM2 has detailed logic for initialization of 3D trajectories (for perspective videos) using keyframe selection.
Note that we use a modified version of ORB-SLAM2, following changes proposed by~\citet{zhou2018stereo} for improved accuracy, and restarting ORB-SLAM2 each time it loses track, therefore generating multiple clips for each video.

\paragraph{Rig-based SfM}
For each such clip, we use its first $k{=}32$ ORB-SLAM2-derived
camera poses as initialization for bundle adjustment.
This bundle adjustment stage recomputes correspondences across (and between)
each sub-video using SIFT feature matching, then performs an incremental SfM procedure similar to COLMAP~\cite{schonberger2016colmap} to estimate poses for the rest of the video, periodically performing a full bundle adjustment.
This careful second pass optimizes camera pose and 3D point geometry using all cube faces.
Further, it uses \textit{rig constraints} to ensure that each set of four cube face views is treated as a rigid collection of cameras with
\begin{enumerate*}[label=(\roman*)]
\item a shared center of projection, and
\item fixed relative orientations.
\end{enumerate*}
Note that we use ORB-SLAM2 and our rig-based 360° bundle adjustment pipeline as opposed to modern tools like VGGT~\cite{wang2025vggt} and MegaSaM~\cite{li2025megasam} since those recent methods are not straightforward to extend to 360° videos and would require retraining.

\paragraph{Output}
This procedure outputs a carefully optimized set of \textit{camera trajectories} for each 360° video, which we treat as the basis for our ground truth, along with a \textit{point cloud} annotated with frame visibility for each sequence.

\subsubsection{Metric scale calibration}
\label{sec:metricscale}
Standard metrics for evaluating camera pose estimates, such as Absolute Trajectory Error (ATE), rely on the RMSE between estimated and ground-truth camera poses.
Hence, the \textit{scale} of ground truth poses is a key factor in such metrics.
Unfortunately, SfM produces camera trajectories with arbitrary scale due to the gauge ambiguity.

To attain interpretable, uniform metrics for camera pose, we therefore \textit{metrically re-scale} our pose estimates.
To perform this rescaling, we start from the four cube-face projected perspective sub-videos from the SfM stage of our pipeline. 
We use a state-of-the-art zero-shot monocular \textit{metric} depth estimation model, Depth Pro~\cite{bochkovskii2024depth}, to predict metric depth for each frame of each sub-video.
We then reproject the visible 3D points calculated by our rig-based SfM stage into each frame and compute their depth ($z$-coordinate) with respect to the corresponding virtual camera. 
The ratio of this depth and the metric monocular depth at each pixel gives us an estimate of the scale factor that maps the SfM reconstruction to metric scale. If a 3D point $p$ is mapped to pixel $u$ with a depth $D$, the scale of the frame $f$ is
\begin{equation}
    \text{S}(f) = \text{Median}_p \frac{D_{u(p)}}{p_z} ~~~~~ p \in {\text{visible in} f}
\end{equation}
For each frame, we take the median of the calculated scales as the frame scale. 
We then calculate the mean $\mu_S$ and standard deviation $\sigma_S$ of the calculated metric frame scales across all frames. 
For robustness, we only consider clips where the standard deviation of frame scales is less than the mean ($\sigma_S < \mu_S)$.
For clips that pass this test, the average of all frame scales $\mu_S$ is accepted as the clip scale, and we apply that scale factor to map the entire clip to our approximately metric scale.

\subsubsection{Ground-truth cross-validation}
\label{sec:gtverify}
To further validate the trajectory estimates from the 360° rig, we \textit{cross-validate} by running ORB-SLAM2 independently on perspective crops around the cube faces, centered at $[90, 180, 270]$ degrees rotation from the frontal view~(we avoid the frontal cube face, as it was used to initialize the 360° estimates).
We employ Umeyama~\cite{umeyama1991least} to align the trajectories estimated from the 360° rig and from the perspective projection runs. 
Given the reference trajectory from the 360° rig, we use standard metrics from the SfM literature~\cite{zhang2018viotutorial}, in particular, Absolute Trajectory Error~(ATE) 
 and Relative Position Error~(RPE), to verify if any estimated perspective trajectory matches the 360° rig trajectory, where ATE and RPE are defined as follows (with separate RPE calculations for translation $t$ and rotation $R$):
 \begin{equation}
    \ATE(g, e) = \Big( \sum_i ||g_i - e_i||_2^2  \Big)^\frac{1}{2}\label{eq:ate}
\end{equation}
\begin{equation}
   \!\! \RPEt(g, e) = \Big( \sum_i \Big|  \frac{||g_i \!-\! g_{i+1}||_2^2 - ||e_i \!-\! e_{i+1}||_2^2 } {||g_i - g_{i+1}||_2^2} \Big| \Big)^\frac{1}{2},\label{eq:rpe_t}
\end{equation}
\begin{equation}
\scriptsize
   \!\! \RPEr(G, E) = 
   \Big( \sum_i \Big|\text{acos}\left(\frac{\text{Tr}((G_{ i}^{-1}G_{i+1})^{-1} (E_{i}^{-1}E_{i+1})) - 1}{2}\right) \Big| \Big)^\frac{1}{2},\label{eq:rpe_r}
\end{equation}
where $i$ denotes a frame index, $g$ and $e$ denote the ground truth and estimated positions, and $G$ and $E$ the rotation matrices, along two camera trajectories in correspondence.
By thresholding these quantities we cull video clips for which our confidence in the estimated camera poses is not high enough to use them as ground truth 
(details in Appendix).

\subsection{Deriving benchmark video clips}
\label{sec:synthesizing}

We obtain benchmark video clips from the full 360° videos by projecting those full videos to perspective crops. The 360° videos allow for a wide range of design choices in producing these crops.  In particular, we can choose any rotational trajectory or field of view (or even dynamic field of view) that we wish.
To produce benchmark videos that mimic the motion of regular cameras (e.g., smartphone or other handheld cameras), we vary the viewpoint direction of our perspective cropped videos across the derived clips, rather than, say, choosing a fixed orientation with respect to the panorama (as the orientation of the panoramic camera itself often doesn't vary much in 360° videos).

If any of the perspective projection estimates fail the verification step in~\Cref{sec:gtverify}, we take that as an indicator of a challenging viewing direction (i.e., ORB-SLAM2 has failed on that cube face). To maximize the difficulty of our test trajectories, we select that cube face as the view for the initial frame. Otherwise, we start from the frontal cube face (i.e., $[0,0,0]$ rotation angles).

Starting from the initial orientation, we vary the virtual camera orientation across the frame (utilizing the 360° captured field of view), where each synthesized trajectory is composed of two components:
(1)~a \emph{low-frequency signal} simulating an observer's intentional
motion (e.g., panning or tilting), and (2)~an optional \emph{high-frequency
noise} component simulating handheld jitter, randomly assigned per clip as
\emph{none}, \emph{medium}, or \emph{large} shake amplitude.
For the low-frequency component, we either retain the original
viewpoint or sample a rotation pattern from one of five
modes: \emph{waypoints}, \emph{scan}, \emph{cinematic}, \emph{roll}, and
\emph{orbit}.
Detailed definitions and parameter ranges are in the supplemental material (Supp.\
Table~\ref{tab:rotation_modes}).

\begin{figure*}[ht]
\includegraphics[width=\linewidth]{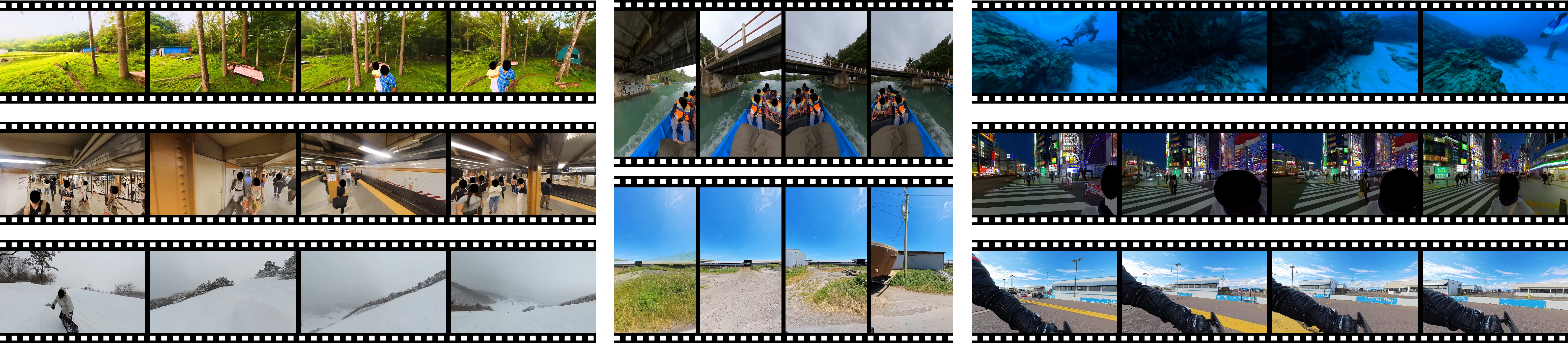}
\vspace{-0.2em}
\captionof{figure}{
    \textbf{Example benchmark videos.}
    Our benchmark video clips show a diversity of scenes with varying degrees of dynamic object presence, reprojected to different rotation patterns, fields of view, and aspect ratios.
}
\label{fig:\currfilebase}
\end{figure*}

The field of view is sampled uniformly from $[30^\circ, 120^\circ]$, where
narrower values simulate telephoto zoom and wider values approximate
ultra-wide-angle lenses. The resulting perspective-projected videos
constitute our ORBIT benchmark (Fig.~\ref{fig:samples}).

These perspective videos inherit ground truth camera poses from the 360° videos (with rotations suitably transformed according to the per-frame perturbations described above).

\subsection{Challenges faced by SfM methods} 
\label{sec:challenge}
Most classic SfM methods rely on stable keypoints or textures that can be tracked across frames. As such, two key types of challenges include (1) a lack of stable keypoints and (2) the presence of outlier or moving keypoints on dynamic objects. 
Most of the clips in our benchmark include one or both of these challenges. 
Nearly all clips ($>70\%$) possess moving objects in all or part of the frames. We further categorize the challenges present in the videos below:

\paragraph{Low texture in static regions} Our benchmark includes videos that are captured in snow, in the desert, and underwater or have some frames that look at concrete, texture-poor walls, and sky. Approximately $12\%$  of our clips are situated in areas without definitive texture or distinctive keypoints. 
Our GT validation technique can often successfully pose and validate such difficult clips because other cube faces may have sufficient texture (mountains alongside sand or forests alongside water), even if such texture is not visible in the test viewpoints.

\paragraph{Dark scenes} 
Dark, low-contrast scenes also hinder SfM algorithms. Our benchmark contains scenes that are captured either at night or in dark areas like caves. Such scenes also constitute approximately $12\%$ of our clips.

\paragraph{High camera speed} Large displacements from one frame to the next can break certain assumptions made in SLAM-based methods. 
Depending on the camera mounting method, high speeds can also lead to jittery videos, which we find to be particularly challenging to methods that rely on smoothness of frame-to-frame motion. 
Some of our clips ($\sim 25\%$) are captured from inside moving vehicles, while paragliding, while skiing, or in otherwise high-velocity scenarios. 
Mistakes in these scenes can also result in higher ATE error values due to larger ground truth trajectory motion.

\paragraph{Camera rotation} Since our clips are selected from 360° videos we have the freedom of selecting camera rotation at each frame. 
To this end we introduce challenging rotations during rendering both as low-frequency structured noise (simulating deliberate camera motions) and uniform noise (simulating jitter) as described in the previous section. We noticed that fully feedforward methods such as VGGT-Long and MonST3R are more prone to error in the presence of large camera rotations.

\paragraph{Crowded scenes} About $16\%$ of our clips are captured in crowded areas such as busy streets and tourist sites. In these cases 
a significant fraction of the field or view of each frame is occluded by moving objects, reducing the visibility of the static and reliably trackable portions of the scene. 
We observe that classic feature-matching-based methods like COLMAP tend to fail more often in such scenarios.

\paragraph{Egocentric Captures} An especially challenging 
scenario occurs when a moving object is in a constant position with respect to camera. 
Such examples are commonplace, e.g., while capturing ego-motion such as a boat or a car that the camera itself is riding on. 
In such cases there can be two frames of reference for the position of camera, one with respect to the world and one with respect to the object that is moving with the camera; 
note that we always choose the world frame of reference as the correct one, i.e., the coordinate frame of the static Earth. 
We observe that training-based methods (MegaSAM and VGGtLong) are more susceptible to this failure mode.

\paragraph{Fluid texture} Large bodies of water are an extreme case of dynamic scenes since the water is in constant motion while having a locally repeatable texture. 
The fluidity of water introduces many noisy matches between one frame and the next which hinders methods that lack a strong robustness criterion.

\subsection{Failure modes}
One critical aspect when comparing the performance of different SfM methods is to account for the possibility of failing to output any camera estimates. 
Some SfM methods, such as MegaSaM and feedforward approaches like VGGT, always output a camera estimate (no matter how inaccurate). 
Other methods, such as COLMAP, sometimes drop a few frames, or can at times completely fail on a clip and produce no output whatsoever,
due, e.g., to an inability to find sufficient reliable feature matches in some or all frames. 
In our evaluations, if some frames are dropped, we copy the previously reported camera pose for the dropped frames until a new estimate is reported.

To quantify the failure rate of various algorithms, beyond ATE (Eq.~\ref{eq:ate}) and RPE (\cref{eq:rpe_t,eq:rpe_r}), we set a threshold on the ATE and RPE for each clip and consider it to have completely failed when the method fails to generate a camera estimate or the errors exceed this threshold (in addition to total failure to produce any output in the case of methods like COLMAP).
The average success rates for each algorithm are reported in the last column of Table~\ref{tab:benchmark}.

\section{Results}
\vspace*{-0.1cm}

We next perform a quantitative evaluation of several leading algorithms and discuss some of the main challenging issues.

\begin{table*}[t]
\centering
\small
\setlength\tabcolsep{2pt}
\begin{tabular}{lccccc}
\toprule
Method & ATE Median  $\downarrow$ & ATE  $\downarrow$ &  RPE-R  $\downarrow$ & RPE-T  $\downarrow$ & Failure Rate  $\% \downarrow$ \\
\midrule
COLMAP & 0.40 & 1.34 $\pm$ 2.16 & 1.19 $\pm$ 2.53 & 0.13 $\pm$ 0.28 & 56.49 \\
MegaSaM & 0.13 & 0.67 $\pm$ 1.24 & 0.44 $\pm$ 0.89 &  \textbf{0.04 $\pm$ 0.10} & 39.09  \\
RoMo+MegaSaM  & \textbf{0.11} & \textbf{0.59 $\pm$ 1.18} & \textbf{0.43 $\pm$ 0.93} & 0.05 $\pm$ 0.12 & \textbf{38.56}\\
ORB-SLAM2 & 1.78 & 2.55 $\pm$ 2.71 & 1.20 $\pm$ 1.70 & 0.13 $\pm$ 0.25 & 95.45 \\
VGGT-Long & 0.84 & 1.30 $\pm$ 1.35 & 1.55 $\pm$ 1.58 & 0.15 $\pm$ 0.18 & 94.72 \\
MonST3R & 1.34 & 1.96 $\pm$ 1.79 & 1.71 $\pm$ 1.90 & 0.22 $\pm$ 0.44 & 98.44 \\
\bottomrule
\end{tabular}
\vspace*{0.1cm}
\caption{\textbf{Evaluating camera pose estimates with ORBIT}. We report the ATE median, as well as the mean and standard deviation of ATE, RPE-T, and RPE-R. Finally, we report failure rate as defined in Sec.~\ref{sec:eval}.}
\label{tab:benchmark}
\end{table*}

\subsection{Ground truth pipeline fidelity}
We expect the fidelity of our pipeline's outputs to be defined by the thresholds used in the cross-validation step. 
We validated the fidelity of our full pipeline on two datasets with known ground truth camera poses:
\begin{enumerate}
    \item \textbf{360Loc~\cite{huang2024360loc}:} A real-world dataset combining LiDAR and 360° camera captures in natural indoor/outdoor environments similar to ORBIT, captured with LIDAR to enable ground truth metric camera poses.
    \item \textbf{Synthetic 360:} A set of photorealistic Blender scenes containing dynamic objects rendered to photorealistic 360° videos.
\end{enumerate}

\vspace*{0.1cm}
\noindent \textbf{Results:} On 360Loc, our pipeline achieves an Absolute Trajectory Error (ATE) of $0.07 \pm 0.04\textrm{m}$ and a Relative Pose Error (RPE-R) of $0.13 \pm 0.07$. 
On the Synthetic 360 dataset we achieve an ATE of $0.0 \pm 0.0\textrm{m}$ (i.e., we very closely match the ground truth) and RPE-R of $0.02 \pm 0.005$. 
These errors are orders of magnitude smaller than the failure cases of baselines (often $>1$m), suggesting that our ORBIT pipeline produces poses of sufficient accuracy to distinguish algorithmic performance.
Furthermore, we employ strict cross-validation thresholds during generation (see Appendix).

\begin{figure}[t]

\centering
\setlength{\tabcolsep}{1.5pt}
{\scriptsize
\begin{tabular}{c c}
    \multicolumn{2}{l}{Original frames} \\
    \includegraphics[width=0.232\textwidth]{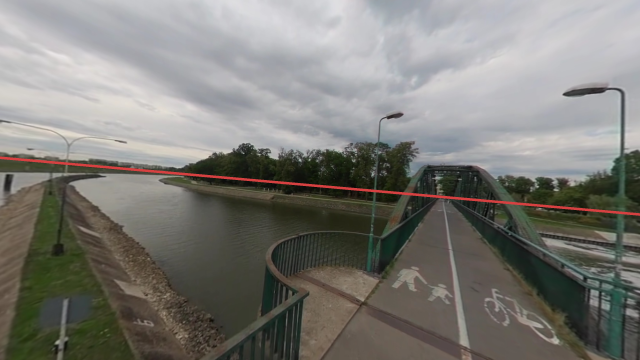} &
    \includegraphics[width=0.232\textwidth]{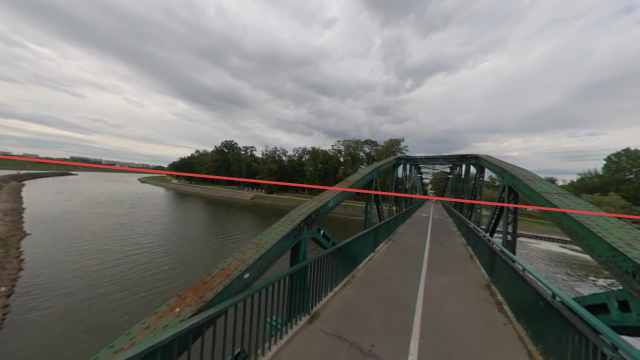} \\
    \multicolumn{2}{l}{Stabilized  + tracked points } \\
    \includegraphics[width=0.232\textwidth]{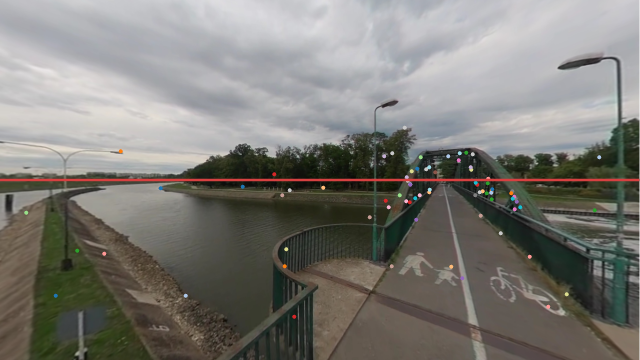} &
    \includegraphics[width=0.232\textwidth]{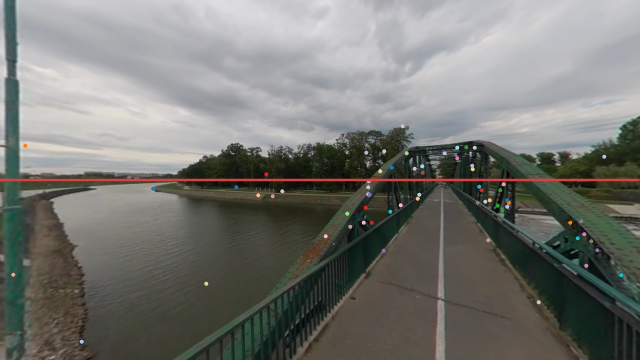} \\
\end{tabular}
}   
\captionof{figure}{
    \textbf{Visual verification.}
    Given our estimated ground truth camera poses, we stabilize the 360° videos. 
    Red lines depict the horizon in the input equirect frame (first row), as well as the equirect following stabilization (second row). 
    We also plot reprojected tracked 3D points via our estimated ground truth trajectories.
    These augmented videos assist us in validating our pseudo-ground truth camera trajectories.
    }
\vspace*{-0.8em}
\label{fig:\currfilebase}
\end{figure}

We also perform a manual  verification of stabilized, oriented clips.
We align the first frame to gravity~\cite{veicht2024geocalib,jin2023perspective}, then warp all subsequent frames to match the orientation of this first frame using our SfM-estimated relative rotation. Fig.~\ref{fig:stabilization} illustrates these steps. 
In the stabilized 360° video, rotation errors cause obvious bending of vertical lines or background drift, which serve as visual cues for human evaluators to reject poor-quality trajectories. 
We further project tracked 3D points onto each frame, and ask human evaluators to verify that points on static regions remain locked to scene content across frames.

\subsection{Evaluations}\label{sec:eval}
We evaluate several methods, including state-of-the-art SfM methods.
The current gold standard in many SfM applications, including novel view synthesis, is COLMAP \cite{schonberger2016colmap} due to its reliability and use of careful incremental bundle adjustment to produce accurate camera poses. 
Therefore, as a strong baseline, we evaluate COLMAP's performance on our benchmark. We also evaluate:
\begin{itemize}[listparindent=0pt,leftmargin=*,itemindent=0pt]
\item ORB-SLAM2 \cite{mur2015orb}, a leading SLAM method based on traditional, handcrafted (fast) feature matching.
It can drop frames if it cannot reliably track them.

\item MonST3R \cite{zhang2024monst3r}, which extends DUSt3R \cite{Wang2024dust3r} to include dynamic scenes by projecting onto 3D point clouds estimated from pairs of frames.

\item MegaSaM \cite{li2025megasam}, a recent state-of-the-art method that has gained traction due to its excellent performance on available benchmarks. 
MegaSaM is based on the DROID-SLAM learned bundle adjustment framework~\cite{Teed2021droidslam}, and utilizes monocular depth estimation, along with predicted motion probabilities, within a robust differentiable bundle adjustment.

\item RoMo \cite{goli2025romo}, a recent method for zero-shot motion segmentation for SFM applications. RoMo produces motion masks that can be used as an input to other SFM methods. Here we combined the masks with MegaSaM as the best available method on our benchmark.

\item VGGT \cite{wang2025vggt}, a leader among the recent family of feedforward methods that train neural network to directly predict camera poses and 3D point maps from a set of input frames. VGGT-Long \cite{deng2025vggt} is a recent method that extends VGGT to clips with more than 200 frames (the limit of VGGT in our tests). Since our benchmark videos often have more than 200 frames, we report the results of VGGT-Long.
\end{itemize}

We report the performance of all of these methods on ORBIT in Table.~\ref{tab:benchmark}. 
This table presents the average and the median Absolute Trajectory Error (ATE), as well as relative errors for both camera translation (RPE-T) and rotation (RPE-R).

The results show that all methods exhibit varying degrees of poor performance on \benchmark. Some techniques including COLMAP and ORB-SLAM2 fail completely on some clips.  Overall, all measurements show a large standard deviation (due to catastrophic failures on a substantial portion of clips), suggesting that average ATE, RPE-T, and RPE-R are not robust for challenging SFM tasks and especially on our benchmark. Hence we also report median ATE and define a failure rate.

\begin{figure}[t]

\begin{center}
\vspace*{-0.2cm} 
\includegraphics[width=0.95\linewidth]{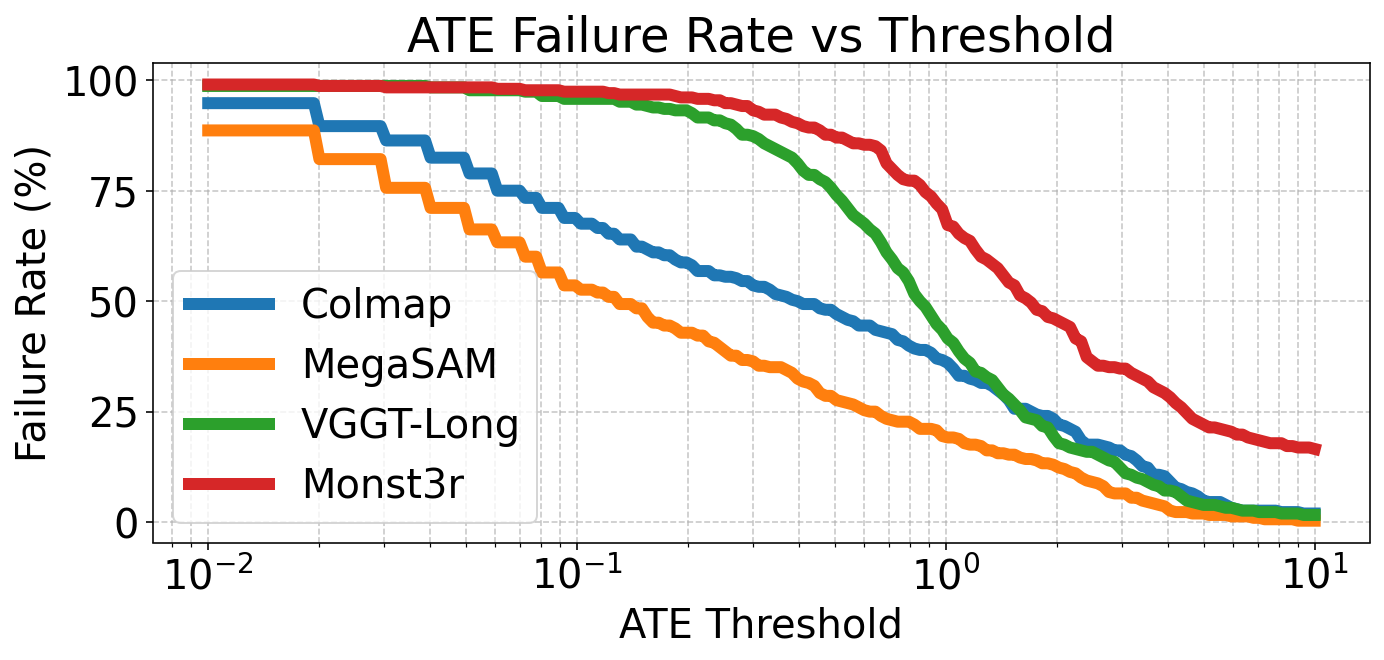}
\end{center}
\vspace*{-0.25cm}
\captionof{figure}{ 
    \textbf{ATE Distribution.}
    Here we sweep a threshold on a failure measure (defined in terms of the ATE being above a given threshold), and plot failure rate as a function of threshold.
    We observe different distributions between fully feedforward methods (VGGT-Long and MonST3R) and optimization-based methods (e.g., COLMAP, MegaSaM w/w.o Romo).
    }
\label{fig:\currfilebase}
\vspace*{-0.2cm}
\end{figure}

For our main failure metric, we define a threshold of $\text{ATE} > 0.4,~\text{RPE-R} > 0.4$ which is at least 2$\times$ larger than the thresholds we used for cross validation in Sec.~\ref{sec:gtverify}. If a method has an ATE and RPE-R that are larger than 40cm or 20° on average, we deem the method to have completely failed and determine that the measurements are only skewing the averages.
While VGGT-Long achieves an average ATE slightly better than that of COLMAP, it has a $94.7\%$ failure rate in the most strict matching criteria. This suggest that COLMAP is still a good choice in less challenging application scenarios.

We further investigate the distribution of clip ATEs for each method by 
sweeping the ATE threshold and computing a failure ratio at each threshold, shown in Fig.~\ref{fig:ate_success}. 
We find that feedforward methods like MonST3R and VGGT-Long rarely match the ground truth trajectory as well as bundle adjustment--based methods such as COLMAP and MegaSaM. Considering the lower std of VGGT-Long in Tab.~\ref{tab:benchmark}, Fig.~\ref{fig:ate_success} show cases a distributional difference between feedforward and bundle-adjustment based methods.

\begin{table*}[t]
\centering
\small
\setlength\tabcolsep{2pt}
\begin{tabular}{lcccccccccccc}
\toprule
Challenge & \multicolumn{2}{c}{COLMAP} & \multicolumn{2}{c}{MegaSaM} & \multicolumn{2}{c}{RoMo} & \multicolumn{2}{c}{ORB-SLAM2} & \multicolumn{2}{c}{VGGT-Long} & \multicolumn{2}{c}{MonST3R} \\
 & ATE & RPE-R & ATE & RPE-R & ATE & RPE-R & ATE & RPE-R & ATE & RPE-R & ATE & RPE-R \\
\midrule

Speed & 0.90 & 1.21 & 0.49 & 0.79 & 0.43 & 0.78 & \textcolor{red}{3.59} & \textcolor{red}{1.41} & \textcolor{red}{1.61} & 2.11 & \textcolor{red}{2.33} & 1.62 \\
Crowd & 0.43 & 1.07 & 0.08 & \textcolor{blue}{0.23} & 0.08 & \textcolor{blue}{0.20} & 1.79 & 1.12 & 0.81 & 1.17 & 1.51 & \textcolor{red}{1.87} \\
Ego & 0.39 & \textcolor{blue}{1.04} & 0.22 & 0.51 & 0.20 & 0.52 & 1.96 & 1.23 & 0.93 & 1.65 & 1.25 & 1.68 \\
Fluid & \textcolor{blue}{0.20} & 1.73 & \textcolor{blue}{0.05} & 0.45 & \textcolor{blue}{0.06} & 0.40 & \textcolor{blue}{0.85} & \textcolor{blue}{1.05} & \textcolor{blue}{0.59} & 1.75 & \textcolor{blue}{0.89} & 1.63 \\
Light & 0.50 & 2.38 & 0.08 & 0.31 & 0.08 & 0.23 & 1.76 & 1.23 & 0.69 & \textcolor{blue}{0.91} & 1.20 & 1.56 \\
Texture & \textcolor{red}{1.59} & \textcolor{red}{2.39} & \textcolor{red}{0.58} & \textcolor{red}{1.04} & \textcolor{red}{0.78} & \textcolor{red}{1.13} & 2.67 & 1.35 & 1.38 & \textcolor{red}{2.42} & 1.30 & \textcolor{blue}{1.32} \\

\bottomrule
\end{tabular}
\vspace*{0.1cm}
\caption{\textbf{Metrics by challenge category}. Median ATE and average RPE-R highlighting easiest (blue) and hardest (red) challenge categories for each method and according to each metric.}
\label{tab:challenge_metrics}
\vspace*{-0.3cm}
\end{table*}

\begin{figure}[t]

\begin{center}
\includegraphics[width=0.95\linewidth]{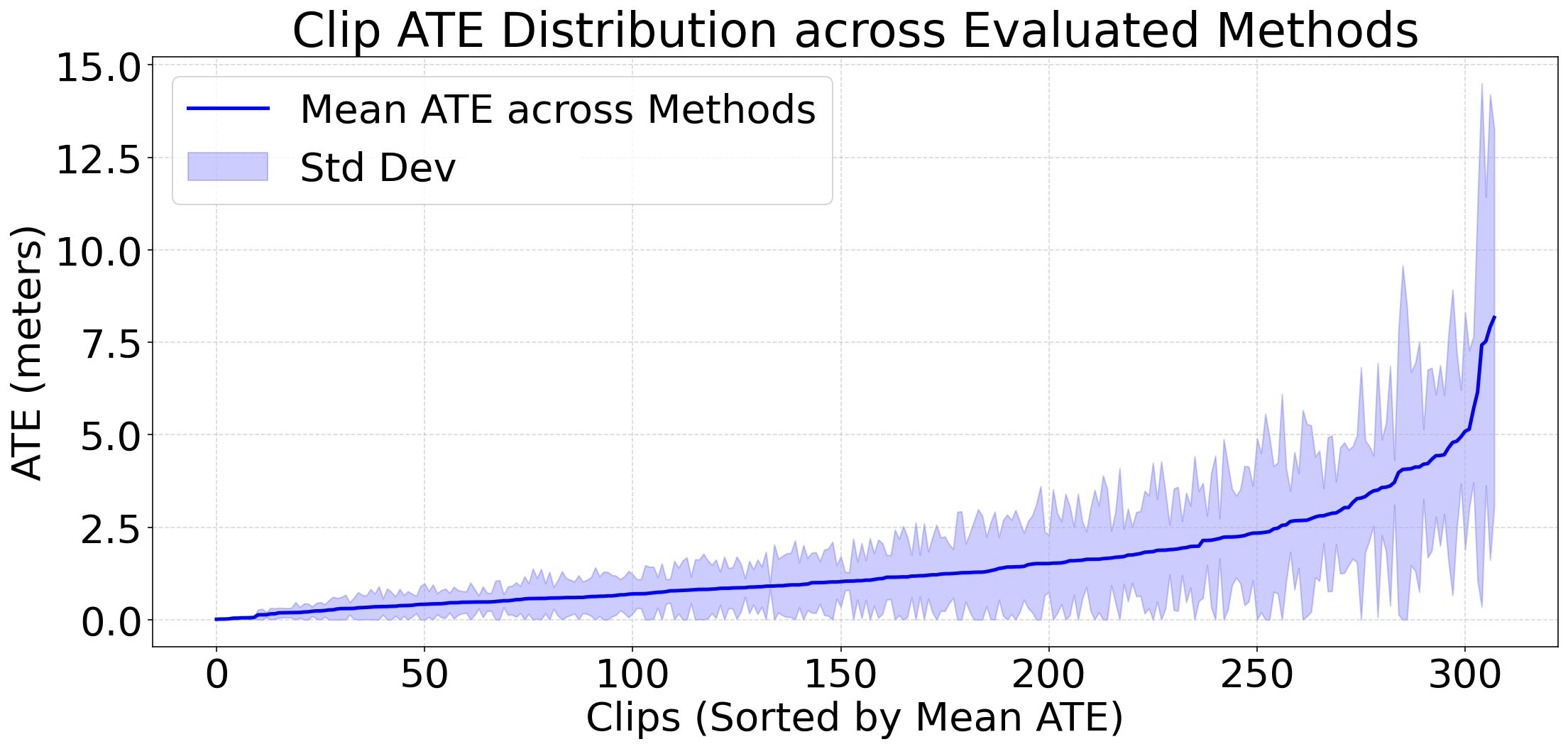}
\end{center}
\vspace*{-0.25cm}
\captionof{figure}{ 
    \textbf{Variance in method performance.}
    We sort each ORBIT clip by average ATE across methods and plot this average and the standard deviation on each clip. A high standard deviation indicates a clip where different methods exhibit highly varying performance. The high variance of method performances observed in this plot agrees with our finding of distinct failure modes per method.    %
    }
\label{fig:\currfilebase}
\vspace*{-0.2cm}
\end{figure}

One of the main advantages of \benchmark is its diversity of challenging cases. As a measure of the diversity of our clips, we calculate the average and standard deviation of performance on each clip across the methods we evaluated. Fig.~\ref{fig:ate_clip_method} shows that many clips have large variations in terms of different method performance. For most 
clips, even ones with high average ATE, there is usually at least one method with a relatively small ATE that successfully estimates the trajectory. Next, we explore particular challenge cases for different methods.

\subsection{Challenge Analysis}

To further analyze the statistics of \benchmark in terms of types of challenge scenarios, we label each clip based on the presence of each of the types of challenges described in Sec.~\ref{sec:challenge},
\ie, high speed, low texture, low light, crowds, ego-centric objects, and the presence of large bodies of water/fluids. 
Table~\ref{tab:challenge_metrics} reports the median ATE and average RPE-R for each method on subsets of clips corresponding to each challenge. 

We observe that lack of texture is a key challenge for bundle adjustment--based methods like COLMAP and MegaSaM. 
Interestingly, adding RoMo masking to MegaSaMs degrades the performance on texture-poor clips while improving performance in all other challenge categories. 
In contrast, 
for feedforward methods like VGGT-Long, 
high-speed camera motion appears to be the most challenging category.
While MegaSaM improves upon COLMAP on egocentric objects, MegaSaM shows major robustness in the crowd category.

\begin{figure}[h]
\begin{center}
\includegraphics[width=.95\linewidth]{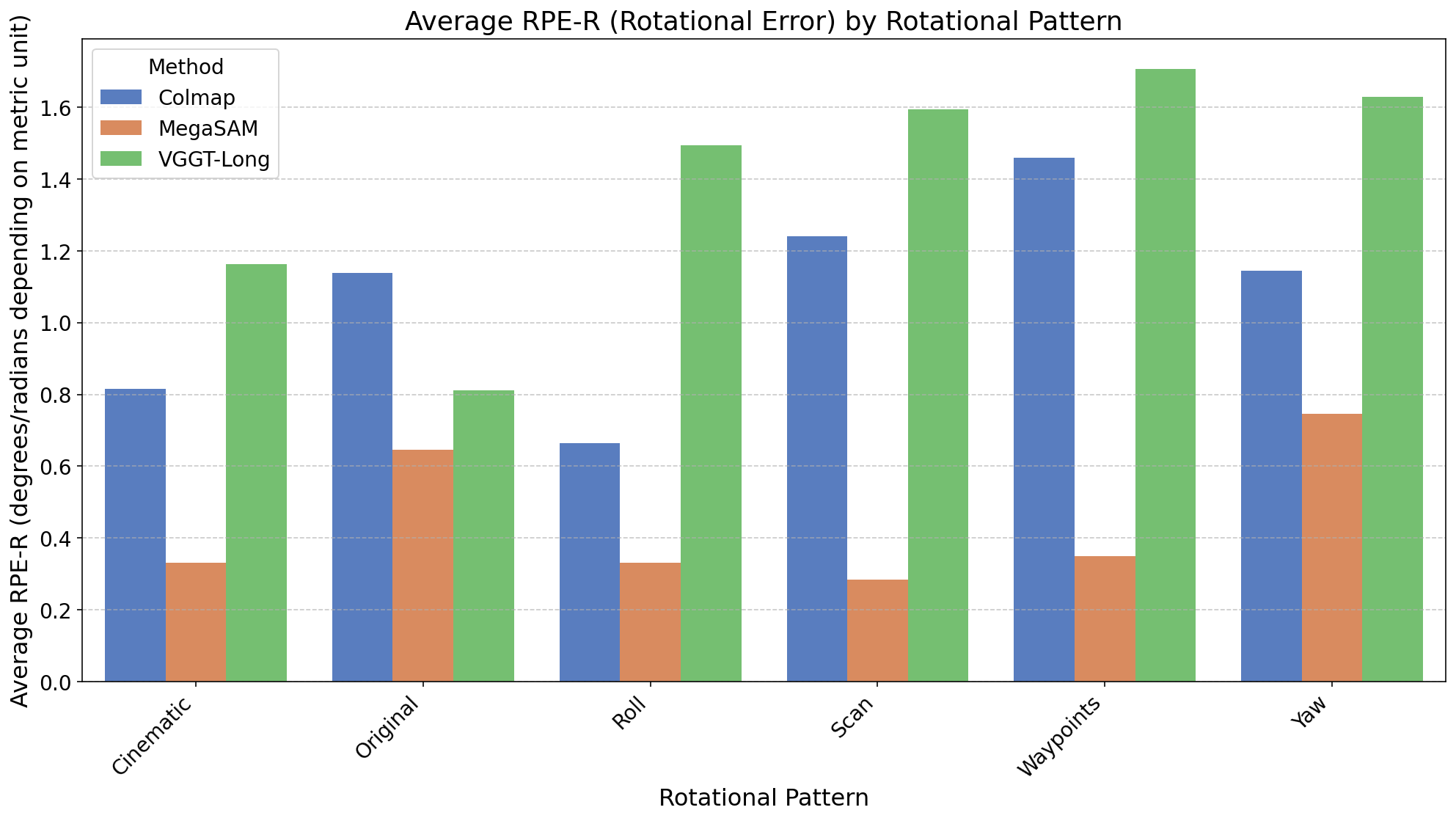}
\end{center}
\vspace{-0.5cm}
\captionof{figure}{
    \textbf{RPE-Rot based on Rotation Pattern in \benchmark.} VGGT-Long shows a strong bias toward performance on original rotation patterns.
}
\label{fig:\currfilebase}
\vspace{-0.6cm}
\end{figure}

\paragraph{Rotation patterns} We analyze the various patterns of rotations we introduce into our derived perspective videos in more detail by comparing per-pattern performance in Fig.~\ref{fig:rotation-rpe}. One noticeable observation is that the feedforward VGGT-Long method underperforms bundle adjustment based methods (MegaSam and Colmap) on all patterns except the original patterns which it outperforms Colmap. This highlights a possible dataset bias during training of the feedforward method that can be mitigated by diversifying the camera rotations while training feedforward methods. 

Overall, Tab.~\ref{tab:challenge_metrics}  and Fig.~\ref{fig:rotation-rpe} show that \benchmark exposes a diverse set of challenges and is a valuable tool for diagnosing and analyzing SfM methods.

\section{Conclusions}
\vspace*{-0.1cm}
The field of camera pose estimation is currently hindered by a lack of challenging and realistic evaluation benchmarks. 
Many recent methods have claimed advantages over older, classic methods such as COLMAP by evaluating on currently available benchmarks, but these benchmarks are largely synthetic or simple. 
This lack of a proper benchmark can potentially lead researchers into misleading directions while not addressing 
issues faced by real-world users. 

We propose a new real-world SfM benchmark, ORBIT, derived from 360° videos, to address this problem.
ORBIT is challengingly diverse, including many indoor and outdoor scenes with moving objects.  We augment the videos with camera trajectories computed using our omnidirectional rig-based SfM pipeline. The camera trajectories themselves are also diverse, with loops as well as straight paths, and include varying camera rotations as well as translations.
Our analysis on current SOTA methods shows that there is a large performance gap 
with ground truth and many challenges are still left open and unanswered. 
We view ORBIT as an evolving benchmark; for instance, in the future, we plan to expand ORBIT to include variations in focal lengths to enable intrinsics prediction challenges.
ORBIT and the related codebase are available at \url{https://orbit-sfm.github.io}.

\vspace*{0.25cm}

\noindent
{\bf Acknowledgements}
We express our sincere gratitude to Charles Herrmann for his  early contributions to the project.

\clearpage
\setcounter{page}{1}
\maketitlesupplementary
\appendix

\section{Method Parameter Details}

\paragraph{Metric Scale Recalibration}
When computing metric scale for each reconstructed sequence in our dataset (Section~\ref{sec:metricscale} of the main paper), for the sake of computational efficiency, we subsample the frames by 10$\times$. 
Furthermore, for each cube face, if there are more than 100 visible points, we only consider the first 100 visible points visible to that face for the purpose of estimating metric scale.
For depth estimation, we project the 360° frame into 90° FOV cube faces, resized to $512\times512$. 
We use the default configuration for Depth Pro~\citep{bochkovskii2024depth} as follows: 
\begin{itemize}
\item patch encoder preset: ``dinov2l16 384''
\item   image encoder preset: ``dinov2l16 384''
\item    decoder features: 256
\item    use fov head: True
\item   fov encoder preset: ``dinov2l16 384''.
\end{itemize}

\paragraph{Ground Truth Cross-Validation}
When cross-validating the data, we use as ground truth (Section~\ref{sec:gtverify}), in order to account for different speeds of motion in different cameras, we define a per clip threshold measurement as the median of translational distances between nearby frames in the ground truth (rig-based) trajectory. 
As such, we compare the ATE not only with a fixed threshold based on meters but also with its median speed moment as well.
We define the per clip threshold $\tau$ as:
\begin{equation}
    \tau(g) = \text{med}(||\mathbf{g}_i - \mathbf{g}_{i+3}||_2^2)
\end{equation}
In order for a rig-based 360° trajectory $\mathbf{g}$ to be considered verified against a per-cube-face trajectory $\mathbf{e}$, the trajectory $\mathbf{g}$ should satisfy all of the following: 
\begin{subequations}
\begin{align}
\mathrm{ATE}(\mathbf{g},\mathbf{e}) &< 2 * \tau(\mathbf{g}) \\
\mathrm{RPE}_r(\mathbf{g},\mathbf{e}) &< 0.2 \\
\mathrm{RPE}^*_t(\mathbf{g}, \mathbf{e}) &< 1.0,
\end{align}
\end{subequations}
where $\mathrm{RPE}^*_t(\mathbf{g},\mathbf{e})$ is a normalized version of Eq.~\ref{eq:rpe_t} where we divide by the corresponding translational differences in $\mathbf{g}$.

\paragraph{Rotation Augmentation}
When synthesizing benchmark videos (Section~\ref{sec:synthesizing} of the main paper), for each clip we either retain the original viewpoint or apply a randomly selected rotation pattern.
The applied rotation for frame $i$ consists of a low-frequency intentional motion component $L_i$ and an optional high-frequency handheld noise component $N_i$:
\begin{equation}
    R_i = N_i \, L_i \, B,
\end{equation}
where $B$ is the initial viewing rotation matrix.

The low-frequency component $L_i$ is generated according to one of five modes, summarized in Table~\ref{tab:rotation_modes}.
The base starting yaw is determined by the initial cube-face direction selected during viewpoint selection (Sec.~\ref{sec:synthesizing}), i.e., the challenging cube face where ORB-SLAM2 failed, or the frontal face otherwise.
The remaining base orientation angles are randomly sampled:
\begin{itemize}
    \item Pitch: Gaussian mixture---70\%: $\mathcal{N}(0^\circ, 5^\circ)$, 30\%: $\mathcal{N}(-15^\circ, 12^\circ)$,
    \item Roll: Student's $t$-distribution with $\mathrm{df}{=}5$, $\mathrm{scale}{=}1.5^\circ$.
\end{itemize}

\begin{table*}[h]
\centering
\small
\setlength\tabcolsep{3.5pt}
\begin{tabular}{llp{11cm}}\toprule
\textbf{Mode} & \textbf{Axes} & \textbf{Description} \\ \midrule
Cinematic & Yaw, Pitch & Interpolation from start to end. Yaw: start $\in U(-30^\circ, 30^\circ)$, end $\in U(-60^\circ, 60^\circ)$. Pitch: 70\%: start/end $\in U(-30^\circ, 30^\circ)$, 30\%: $\in U(-60^\circ, 60^\circ)$.
\\Scan & Yaw or Pitch & Sinusoidal horizontal or vertical sweep. Amplitude $\in U(10^\circ, 40^\circ)$, frequency $\in U(0.05, 0.3)$\,Hz.
\\Waypoints & Yaw, Pitch & Interpolation through 3+ randomly placed waypoints. Per waypoint: yaw $\in U(-180^\circ, 180^\circ)$; pitch 70\%: $U(-30^\circ, 30^\circ)$, 30\%: $U(-60^\circ, 60^\circ)$.
\\Roll & Roll & Barrel-roll ($60$--$360^\circ$).
\\Orbit & Yaw & Spin along the vertical direction.
\\\bottomrule
\end{tabular}
\vspace{2mm}
\caption{\textbf{Rotation modes for benchmark synthesis.} For each clip, a mode is randomly selected and its parameters are sampled from the listed ranges. $U(a,b)$ denotes uniform sampling.}
\label{tab:rotation_modes}
\end{table*}

\paragraph{Handheld Noise Simulation}
The high-frequency noise component $N_i$ simulates handheld camera micro-tremors.
We generate Gaussian white noise, temporally smooth it with a 1D Gaussian filter ($\sigma{=}5$ frames), and scale to a target amplitude.
Three noise levels are randomly assigned per clip:
\emph{none} ($0^\circ$), \emph{medium} ($1^\circ$), and \emph{large} ($3^\circ$).
The noise is applied independently to all three Euler angle channels (yaw, pitch, roll).

\begin{figure}[h]
\begin{center}
\includegraphics[width=.85\linewidth]{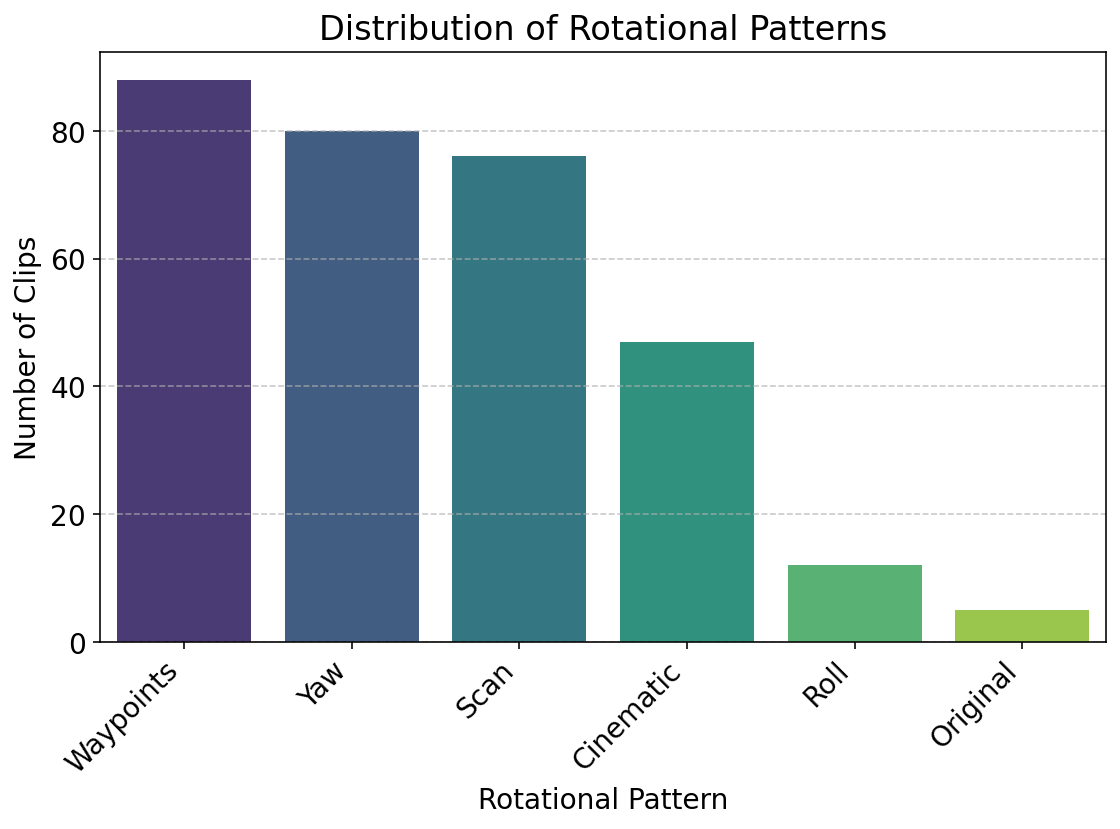}
\end{center}
\vspace{-0.2cm}
\captionof{figure}{
    \textbf{\benchmark Rotation Pattern Distribution --}
    The number of clips in the benchmark for each pattern.
}
\label{fig:\currfilebase}
\vspace{-0.5cm}
\end{figure}

The distribution of patterns in
our benchmark is shown in Fig.~\ref{fig:rotation_count}.

\paragraph{Field-of-View Sampling}
The horizontal field of view (HFOV) for each clip is sampled from $[30^\circ, 120^\circ]$.

\section{Benchmark Method Configurations}
For the sake of aligning the timestamps between ground truth and trajectories if some of the frames are dropped, we use the last previously estimated camera position for the dropped frames. 

\paragraph{COLMAP~\citep{schonberger2016colmap}}
We set the COLMAP settings as follows:
\begin{itemize}
  \item Matcher method: EXHAUSTIVE MATCHER
  
  \item Target percent images initial SfM pass: 1.0
  \item Min images initial SfM pass: 500
  \item Max images initial SfM pass: 1024

  \item Mapper bundle adjustment global function tolerance: $1e$-$6$
  \item Bundle Adjustment function tolerance: $1e$-$6$
  \item Camera model: SIMPLE RADIAL
\end{itemize}

\paragraph{MegaSaM~\citep{li2025megasam}}
We use Depth Pro~\cite{bochkovskii2024depth} for monocular depth estimation, as well as relative depth for sky prediction and metric depth aligning. We use the default setting for all other hyperparameters.

\paragraph{RoMo~\citep{goli2025romo}+MegaSaM~\citep{li2025megasam}}
We first run RoMo~\citep{goli2025romo} with its default settings, with two iterations using SAM 2~\citep{Ravi2024samv2} features as input as well as RAFT~\citep{raft2020} optical flow estimates and a final SAM 2~\citep{Ravi2024samv2} refinement. We use a batch size of 16 and a learning rate of $2e$-$2$ and a ratio threshold of $0.5$ for finding trusted frames. 

\paragraph{MonST3R~\citep{zhang2024monst3r}} In contrast to multi-frame feedforward methods (\eg, VGGT~\citep{wang2025vggt} and VGGT-Long~\citep{deng2025vggt}), MonST3R is designed for two-frame inputs rather than sequences exceeding a thousand frames.
To adapt the model for longer sequences, we first resize the input images such that the shorter dimension is 224 pixels.
We then employ window-based inference with a window size of 50 and an overlap ratio of 0.3.
For all remaining hyperparameters, we follow to the original configuration.

\paragraph{VGGT-Long~\citep{deng2025vggt}}
We use the default base config of VGGT-Long in their official release GitHub repository with a chunk size of 60, overlap of 30, and loop chunk size of 20. This method uses dense alignment, and we use the following IRLS parameters of $\delta = 0.1$, $\text{max\_iters} = 5$, and $\text{tolerance} = 1e$-$9$.

\paragraph{ORB-SLAM2~\citep{mur2015orb}}
We use the default settings of ORB-SLAM2 with a maximum of 4,000 features per frame and a camera scale of 1.0. 
A sequence is counted as a failure if fewer than 30 frames were tracked or if 2 consecutive frames were dropped.
As a result, ORB-SLAM2 only outputs camera estimations for 65 clips and fails on the rest.

\section{Trajectory Comparisons}
Please see the accompanying video clips on the provided webpage for better visualizations of our test clips and of the qualitative performance of each benchmark method. 
Fig.~\ref{fig:traj_h0} and Fig.~\ref{fig:traj_v4} show sample frames and trajectory graphs comparing the ground truth trajectory and each method's output on cips from ORBIT.
We observe in these trajectory comparisons that feed-forward methods like MonST3R and VGGT-Long almost never produce a perfectly matching estimate. On the other hand, bundle adjustment-based methods like COLMAP and MegaSaM are a perfect match for many frames but they can go completely out of bounds on other frames.
\begin{figure*}[ht]
\vspace*{-.5em}
\includegraphics[width=\linewidth]{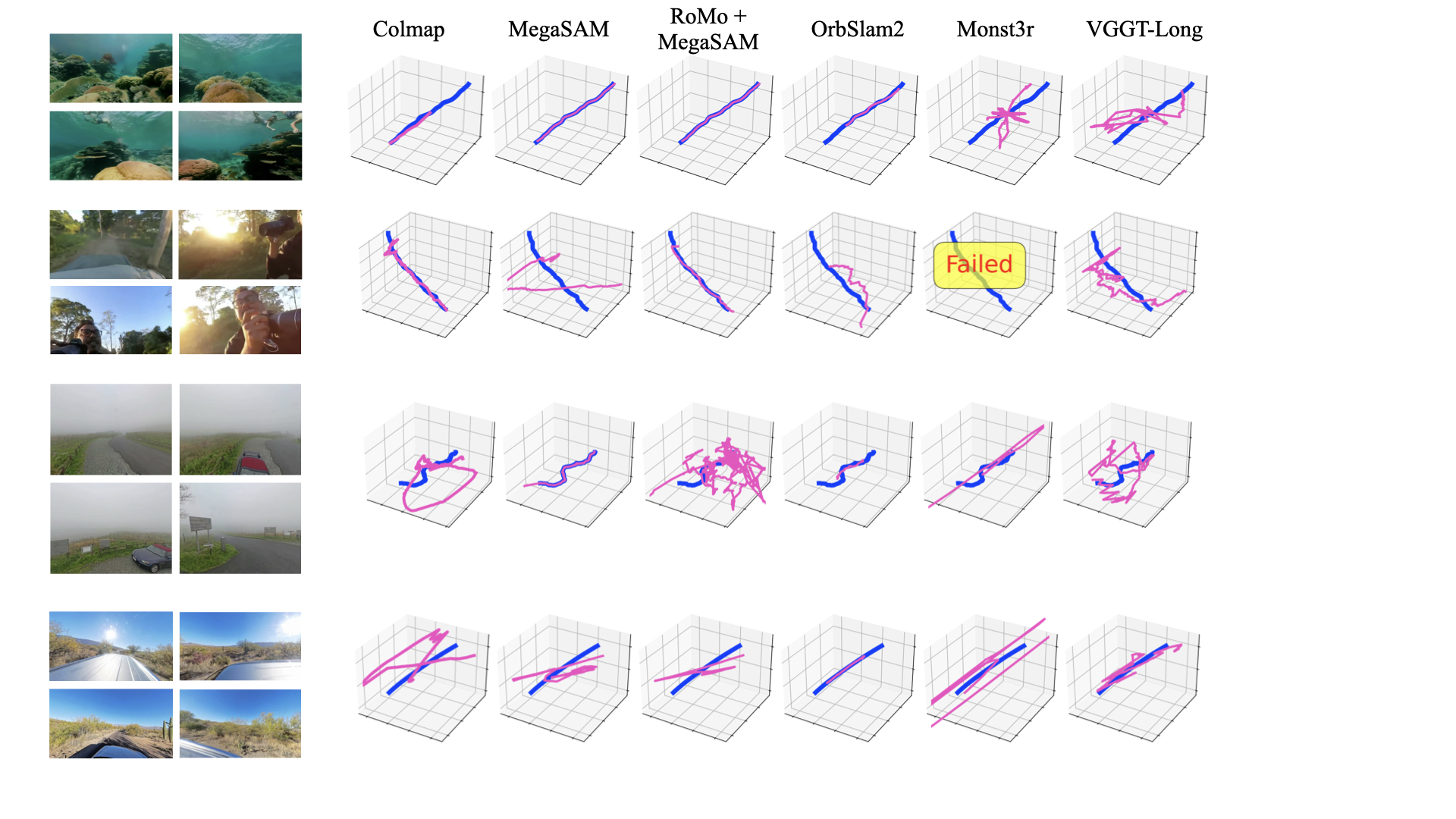}
\vspace*{-.5em}
\captionof{figure}{
\textbf{Trajectory Comparisons.}
Sample frames from \benchmark clips alongside trajectory comparisons between ground truth and each method's estimated trajectory. Ground truth is shown in blue.
}
\label{fig:\currfilebase}
\end{figure*}

\begin{figure*}[ht]
\vspace*{-.5em}
\includegraphics[width=\linewidth]{assets/\currfilebase.png}
\vspace*{-.5em}
\captionof{figure}{
\textbf{Trajectory Comparisons.}
Sample frames from \benchmark clips alongside trajectory comparisons between ground truth and each method's estimated trajectory. Ground truth is shown in blue.
}
\label{fig:\currfilebase}
\end{figure*}

\clearpage

{
    \small
    \bibliographystyle{ieeenat_fullname}
    \bibliography{main}
}

\end{document}